\documentclass[runningheads]{llncs}
\usepackage{amsmath}
\usepackage[T1]{fontenc}
\usepackage{lineno}
\usepackage{graphicx}
\usepackage{hyperref}
\usepackage{color}

\usepackage{listings}
\usepackage{xcolor}
\usepackage{fancyvrb}
\usepackage{caption}
\usepackage{float}
\usepackage{makecell}

\newcommand{\role}[1]{\textcolor{olive}{#1}}
\newcommand{\purpose}[1]{\textcolor{brown}{#1}}
\newcommand{\goalcq}[1]{\textcolor{teal}{#1}}
\newcommand{\task}[1]{\textcolor{violet}{#1}}

\begin{document}
\title{A RAG Approach for Generating Competency Questions in Ontology Engineering}
% \author{Anonymous}
\author{
Xueli Pan\inst{1}\orcidID{0000-0002-3736-7047} \and
Jacco van Ossenbruggen\inst{1}\orcidID{0000-0002-7748-4715} \and
Victor de Boer\inst{1}\orcidID{0000-0001-9079-039X} \and
Zhisheng Huang\inst{1}\orcidID{0000-0003-3794-9829}
}
\authorrunning{X. Pan et al.}
\institute{Vrije Universiteit Amsterdam, De Boelelaan 1105, 1081 HV Amsterdam, Netherlands
\email{\{x.pan2, jacco.van.ossenbruggen, v.de.boer, z.huang\}@vu.nl}}

\maketitle              % typeset the header of the contribution
\begin{abstract}
%Ontology engineering and ontology evaluation have witnessed significant advancements driven by the demand of structured knowledge representation. 
Competency question (CQ) formulation is central to several ontology development and evaluation methodologies. Traditionally, the task of crafting these competency questions heavily relies on the effort of domain experts and knowledge engineers which is often time-consuming and labor-intensive. With the emergence of Large Language Models (LLMs), there arises the possibility to automate and enhance this process. Unlike other similar works which use existing ontologies or knowledge graphs as input to LLMs, we present a retrieval-augmented generation (RAG) approach that uses LLMs for the automatic generation of CQs given a set of scientific papers considered to be a domain knowledge base. We investigate its performance and specifically, we study the impact of different number of papers to the RAG and different temperature setting of the LLM. We conduct experiments using GPT-4 on two domain ontology engineering tasks and compare results against ground-truth CQs constructed by domain experts. Empirical assessments on the results, utilizing evaluation metrics (precision and consistency), reveal that compared to zero-shot prompting, adding relevant domain knowledge to the RAG improves the performance of LLMs on generating CQs for concrete ontology engineering tasks. 

\end{abstract}
%
%
%
%\linenumbers
\section{Introduction}
An ontology is a formal, explicit specification of a shared conceptualization of domain knowledge that can be communicated between humans and computers \cite{1998knowledgeengineering}.
In recent years, significant progress has been achieved in the field of knowledge and ontology engineering due to the surge of data-intensive applications and the growing need for structured knowledge representation. 
Central to many methodologies for the development and evaluation of ontologies are competency questions (CQs), a set of queries in the form of questions that outlining and constraining the scope of knowledge represented in an ontology which an ontology must be able to answer \cite{2019AnalysisofOntology}.
Informal CQs are expressed in natural languages, whereas formal CQs are expressed in the formal language of the ontology \cite{Gruninger1995MethodologyFT}. In this paper,  we focus on informal CQs.

Traditionally, the task of crafting CQs has relied heavily on manual effort of domain experts and knowledge engineers, which is often difficult and time-consuming. 
However, with the emergence of Large Language Models (LLMs), such as OpenAI's GPT-4 and Google's Gemini 1.5, there arises an opportunity to automate and enhance this process.
LLMs are trained on vast amounts of text data and excel at generating human-like text in response to prompts, making them promising tools for aiding the formulation and refinement of CQs.

Recent work has investigated the use of different prompt templates in LLMs for the automatic generation of CQs given an existing ontology \cite{rebboud2024can}.  
However, these methods which generate CQs from existing ontologies or knowledge graphs (KGs) are based on the premise that there is already an ontology or a KG, which is not often the case.
Furthermore, CQs generated from existing ontology might be answerable for the same ontology, which makes these CQs not suitable for ontology evaluation.
In addition, LLMs face the problem of hallucination and low ability to access and manipulate up-to-date knowledge \cite{barnett2024seven}.

Retrieval-augmented generation (RAG)~\cite{lewis2020retrieval} has been introduced to address the above mentioned limitation of LLMs and provide access to domain knowledge, which leads us to investigate to what extent we can use %ask the question about the capability of 
RAG with LLMs to generate competency questions. 

Our contributions are shaped by addressing the following research questions.
\begin{description}
    \item[RQ1:] How well does a RAG-based approach  work in the task of generating CQs given a domain knowledge base, compared to zero-shot prompting?
    \item[RQ2:] How do different parameter settings in a RAG pipeline affect the performance on the task?
\end{description}
To address RQ1, we applied a RAG-based approach to generate CQs for two domain ontology engineering tasks and compared the results to zero-shot prompting. 
For RQ2, we investigated the effects of adjusting two hyperparameters: the number of papers ($N_{paper}$) in the knowledge base and the temperature ($temp$) of the LLMs, on the task performance.
Experimental results reveal that the RAG approach works well in the task of generating CQs for concrete ontologies/KGs that required more domain knowledge.
Moreover, increasing $N_{paper}$ in the RAG pipeline generally improves the performance on the tasks.

% In summary, the contribution of this paper are the following:....
% The remainder of this paper is organized as follows. In Section \ref{sec:related-work} we review some related work. In Section \ref{sec:methodology} we present our RAG approach. In Section \ref{sec:experiment}, we provide details about the two selected ontology engineering tasks and evaluation metrics used in our experiments. In Section \ref{sec:results}, we present the experimental results and the discussion. We conclude and outline some future work in Section \ref{sec:conclusion}.

\section{Related Work}
\label{sec:related-work}

\subsection{Competency questions formalisation}
Traditionally, ontology engineering relies on manual efforts of domain experts and knowledge engineers, especially for the formalisation of CQs.
Different works have been investigated in identifying CQ patterns to improve the automation of CQ formalisation.
Wiśniewski et al. \cite{wisniewski2019analysis} identified 106 CQ patterns by analysing a dataset of 234 CQs and their SPARQL-OWL queries for several ontologies in different domains.
In their follow-up work \cite{wisniewski2021bigcq}, they released BigCQ, the largest dataset of CQ templates with their formalisation into SPARQL-OWL query templates for ontology engineers to use for particular needs.
Based on this work \cite{wisniewski2019analysis}, C. Maria Keet et al. \cite{keet2019claro} designed a template-based controlled natural language (CNL) to author CQs.
In our RAG approach with LLMs, we employ zero-shot prompting and no CQ templates or CQ examples are required for the prompt.

\subsection{Automation of generating CQs}
Several efforts have been investigated on leveraging natural language processing (NLP) or LLMs to generate CQs. %\cite{antia2023automating,alharbirole2024,ciroku2024revont,rebboud2024can}.
The authors of~\cite{antia2023automating} proposed AgOCQs, a method that took advantage of the combination of NLP techniques and a text corpus with CQ templates to generate CQs, whereas no CQ templates are required for our approach.
Rebboud et al.~\cite{rebboud2024can} investigated the suitability of LLMs to automatically generate CQs given an existing ontology, and experiments were conducted with six LLMs and five ontologies.
Alharbi et al.~\cite{alharbi2024experiment} proposed RETROFIT-CQs, a method to generate CQs using LLMs by extracting triples from existing ontologies and feeding them to three prompt templates of an LLM.
These two approaches are based on the premise that there is already an ontology or a KG, which is not often the case.
Our approach addresses this limitation by using the domain literature as input to our RAG pipeline to provide LLMs with accessible and up-to-date domain knowledge.

\section{Methodology}
\label{sec:methodology}
The primary objective of this paper is to investigate the effectiveness of RAG with LLMs to generate CQs in ontology engineering. 
Our RAG approach consists of two components, a RAG pipeline and a prompt engineering.
%we specify our prompt template with crucial parameters. Second, during the generation phase, we implement the RAG pipeline to generate CQs.

\subsection{RAG pipeline}
\label{subsec:rag-pipeline}
\begin{figure}[h]
    \centering
    \includegraphics[width=1.0\textwidth]{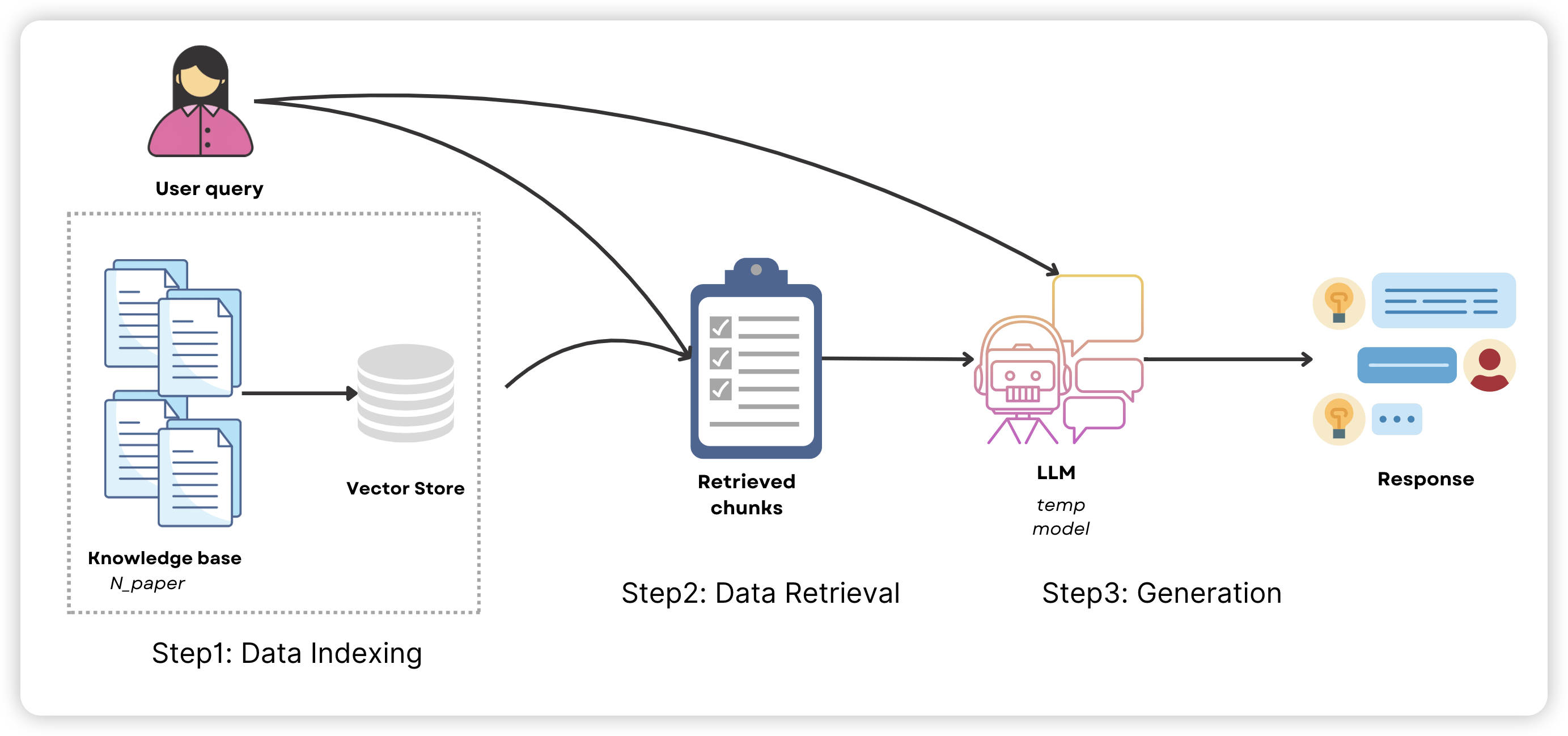}
    \caption{RAG pipeline with three hyperparameters: $N_{paper}, temp$ and $model$}
    \label{fig:rag}
\end{figure}

Figure \ref{fig:rag} illustrates the main steps in the RAG pipeline: domain knowledge indexing, relevant data retrieval, and response generation.

In Step 1, the knowledge base consists of a set of documents that serve as external sources of knowledge to augment the generative ability of LLMs for specific tasks.
During the data indexing process, each selected document is first split into small chunks according to the chunk size defined by different LLMs.
The chunks are then converted to embedding vectors using an embedding model.
The original chunks and their respective vectors would be indexed and stored in a vector database for retrieval.

In Step 2, a user query or a user prompt expressed in natural language is converted to an embedding vector to retrieve relevant chunks from the knowledge base.
Top-k similar chunks are retrieved using a similarity method such as cosine similarity.

In Step 3, the retrieved chunks serve as a context for LLMs to generate the answer for the user query.

In our approach, the documents selected for the knowledge base in Step 1 are scientific papers that inherit domain knowledge for ontology engineering.
One of the pain points in RAG systems is the missing content in the knowledge base \cite{barnett2024seven}, which is also a key challenge in our RAG approach: how to select scientific papers relevant to generate CQs for a specific domain ontology.
There are three principals for selecting scientific papers in our approach: 1) these papers are relevant to the domain of the target ontology; 2) these papers help to explicitly define the scope and requirements of the ontology; 3) the format of the content of these papers should be processable in the RAG system.

To gain a deeper understanding of the effect of papers in the knowledge base in our RAG approach, we set the number of papers ($N_{paper}$) as a hyperparameter to evaluate its impact on the performance of the RAG approach to generate CQs.
Our RAG approach includes two additional hyperparameters: temperature ($temp$) and the model name of the selected LLM ($model$). 
The temperature parameter ranging from 0 to 2 is used to control the randomness of LLMs' output.
There are various open-source LLMs, such as Google's Gemma, as well as closed-source LLMs, like OpenAI's GPT, available for implementing our RAG approach.
For experiments, we choose gpt-4o since it is one of the most advanced LLMs.

\subsection{Prompt engineering}
\label{sec:prompt-engineering}
In the field of LLMs, a prompt serves as an input that directs the model's generation of responses.
The use of prompting techniques involves carefully formulating these prompts to optimize the effectiveness of LLMs.
This process requires intentional structuring and wording of prompts to correspond with the model's abilities and the task description.
Prompts significantly affect the performance of LLMs especially with zero-shot prompting~\cite{jin2022goodpromptworthmillions}. 
Since LLMs are sensitive to prompt, designing a good prompt is another key challenge in our approach. 
For the task of generating CQs, we use zero-shot prompting and consider that a good prompt template in our RAG approach should consist of the following five components.
\begin{itemize}
    \item \textbf{The role of the LLM} In our approach, the LLM is considered as an expert in a specific domain.
    \item \textbf{The purpose of the ontology/KG} The purpose of the ontology/KG is important for LLMs to understand the requirements and scope that the CQs should be covered.
    \item \textbf{The definition of CQ} CQ is a technical term in ontology engineering with a special meaning. Explicitly definition of the CQ in the prompt helps LLMs to better understand the task. 
    \item \textbf{The task description} The number of generative CQs should be determined for each ontology engineering task.
    \item \textbf{The output format} Specifying the structure of the output is beneficial for the analysis and evaluation of the generated output.
\end{itemize}

These components constitute the following template with four variables to be determined for each ontology engineering task in our approach.
Variables are enclosed in curly braces as shown in Listing \ref{lst:prompt-template}.
\vspace{10pt}
\begin{lstlisting}[caption={Prompt template with four variables}, label={lst:prompt-template}]
You are an expert in (*@\role{\{domain name}\}@*).
Your purpose is to (*@\purpose{\{purpose of the ontology}\}@*).
A competency question is (*@\goalcq{\{definition of CQs}\}@*). 
Derive (*@\task{\{number of CQs}\}@*) competency questions for the above-mentioned ontology (or knowledge graph), using the provided documents.
Return ONLY the competency questions, no other text.
\end{lstlisting}
\vspace{-5pt}

\section{Experiment}
\label{sec:experiment}
To evaluate our RAG approach from Section \ref{sec:methodology}, we conduct two experiments, where we 
%We try to 
replicate two domain expert-driven ontology engineering tasks. We choose this setup since these existing experiments provide us with a set of ground-truth CQs identified manually by experts to compare our results to. % based on the 
%We first provide details of the datasets and prompts setting for these two domains, then we describe the evaluation metrics used to evaluate the quality of the CQs generated and the consistency of the output of LLMs. 
%Experiments with each setting iterate for 10 times.
Details about the datasets and codes could be found in this repository \footnote{\url{https://anonymous.4open.science/r/GenCQs/README.md}}.

\subsection{Task description} %ontology/task description
We take two ontology engineering tasks to evaluate our approach for generating CQs.
%ontologies or knowledge graphs with a set of CQs designed by domain experts that could be considered ground truth. 
The first is the construction of a knowledge graph of empirical research in requirement engineering(RE), namely KG-EmpiRE~\cite{karras2023divideconquerempirecommunitymaintainable}.
In this paper, the authors constructed the KG-EmpiRE with the purpose of providing the community with the state and evolution of empirical research in RE.
The KG-EmpiRE was evaluated against 77 CQs manually derived by three domain experts from a published visionary paper about how researchers should conduct empirical research in RE~\cite{sjoberg2007future}.
The second is the construction of a core reference ontology in Human–Computer Interaction (HCI), namely HCIO~\cite{HCIO}.
In this paper, the authors constructed the HCIO with the purpose of clarifying the main concepts involved in the HCI phenomenon.
The HCIO was evaluated against 15 CQs identified by ontology engineers using the methods described in SABiO~\cite{Falbo2014SABiOSA}.
Examples of these CQs identified by domain experts could be found in Table \ref{table:cqs} with column name $CQ_{gt}$.
% summary of the two papers.

\subsection{Hyperparameters setting}
As mentioned in Section \ref{subsec:rag-pipeline}, the selection of documents for the knowledge base in our RAG pipeline is very important.

For the KG-EmpiRE, we follow the methodology of how the authors of KG-EmpiRE derived the ground-truth CQs. 
Since the purpose of KG-EmpiRE is to capture the state and evolution of empirical research in requirement engineering, the authors select a visionary paper \cite{sjoberg2007future} to identify 77 CQs. 
We also select this visionary paper as one of the most important documents in the knowledge base.
To investigate the impact of number of papers in our RAG approach, we also include other related publications on the state and evolution of the topic mentioned in the KG-EmpiRE paper.

For HCIO, since the authors of HCIO do not explicitly mention how these 15 CQs are identified, we select the referenced papers mentioned in the HCIO papers \cite{HCIO} that describe how other relevant ontologies are developed. 

All selected referenced papers are in PDF formats. 
The hyperparameter $N_{paper}$ was set to 1, 2, 3, 4, 5 and 10.
Since gpt-4o is the latest model of OpenAI's GPT models, we set the hyperparameter $model$ to gpt-4o.
Temperature $temp$ is set to 0.5, 0.75, 1.0, 1.25 and 1.5. 
Each experiment, for every hyperparameter configuration, was repeated 10 times.

\subsection{Variable setting for prompts}
We design prompts for the two tasks based on the prompt template described in Section \ref{sec:prompt-engineering}.
This template includes four variables that need to be determined.
The \textit{domain name} is chosen based on the selected task.
The \textit{number of CQs} matches the number of ground-truth CQs provided by domain experts.
For the \textit{definition of CQs}, we take the definition of an informal CQ in \cite{uschold1996ontologies}, a competency question is a natural language question that specifies the requirements of an ontology and can be answered by that ontology.
For the \textit{purpose of the ontology}, we refer to the original papers of KG-EmpiRE \cite{karras2023divideconquerempirecommunitymaintainable} and HCIO \cite{HCIO}, which elaborate the objectives of the corresponding ontology or knowledge graph.

Listing \ref{lst:prompt-re} and Listing \ref{lst:prompt-hci} present the prompts used to generate the CQs for KG-EmpiRE and HCIO, respectively.

% basicstyle=\footnotesize
\vspace{10pt}
\begin{lstlisting}[
%basicstyle=\footnotesize, 
caption={Prompt for KG-EmpiRE}, label={lst:prompt-re}]
You are an expert in (*@\role{Requirements Engineering}@*).
Your purpose is to (*@\purpose{organize scientific data in an openly available and long-term way with respect to building, publishing, and evaluating an initial knowledge graph of empirical research in Requirement Engineering. To achieve this goal, you need to create a knowledge graph which enables sustainable literature reviews to synthesize a comprehensive, up-to-date, and long-term available overview of the state and evolution of empirical research in Requirement Engineering}@*).
A competency questions is (*@\goalcq{a natural language question that specifies the requirements of an ontology and can be answered by that ontology}@*). 
Derive (*@\task{77}@*) competency questions for the above mentioned ontology (or knowledge graph), using the provided documents.
Return ONLY the competency questions, no other text.
\end{lstlisting}

\vspace{10pt}
\begin{lstlisting}[caption={Prompt for HCIO}, label={lst:prompt-hci}]
You are an expert in (*@\role{Human-Computer Interaction}@*).
Your purpose is to (*@\purpose{develop a referece ontology about the human–computer interaction phenomenon. This ontology is grounded in Unified Foundation Ontology and reuses concepts from the core System and Software Ontology to represent the very high-level core concepts in the Human-Computer Interaction and serve as a reference to the HCI domain, with the aim of making a clear and precise definition of domain concepts for the purpose of communication, learning and problem-solving.}@*)
Competency questions are (*@\goalcq{a natural language question that specifies the requirements of an ontology and can be answered by that ontology}@*).
Derive (*@\task{15}@*) competency questions for the above mentioned ontology (or knowledge graph), using the provided documents.
Return ONLY the competency questions, no other text.
\end{lstlisting}
\vspace{-20pt}
\subsection{Evaluation}
We use precision to evaluate the quality of LLM-generated CQs against a set of ground-truth CQs designed by domain experts. 
In addition, we use consistency to evaluate the impact of different temperature settings for the task.

We compared the generative CQs ($CQ_{gen}$) of the LLM to each CQ in the ground truth ($CQ_{gt}$) and consider a $CQ_{gen}$ as valid if it is sufficiently similar to at least one $CQ_{gt}$. 
For the similarity score, we use cosine similarity between the embedding of $CQ_{gen}$ and $CQ_{gt}$ calculated using SentenceBERT~\cite{reimers2019sentence}. 
Similar to~\cite{rebboud2024can}, we define a threshold $\theta$ above which we consider a $CQ_{gen}$ to be valid.
Here we choose the same $\theta$(0.6) as in~\cite{rebboud2024can}.

\vspace{-6mm}

\begin{table}[]
\caption{Examples of generative CQs and their matched ground truth CQ with highest cosine similarity scores for two domain ontology engineering tasks}\label{table:cqs}
\begin{tabular}{|c|l|l|c|c|}
\hline
Domain & \multicolumn{1}{c|}{$CQ_{gen}$} & \multicolumn{1}{c|}{$CQ_{gt}$} & cos & valid $CQ_{gen}$ \\ \hline
hci & \makecell[l]{What are the primary \\ components of a Human-\\Computer Interaction system?} & \makecell[l]{What does make up the \\ user interface of an \\ interactive computer system?} & 0.7393 & yes \\ \hline
hci & \makecell[l]{What methodologies exist \\ for building ontologies \\ in the HCI domain?} & \makecell[l]{What is a complex interactive \\ computer system?} & 0.3467 & no \\ \hline
re & \makecell[l]{How has the use of \\ empirical methods in RE \\ evolved over time?} & \makecell[l]{How often are which \\ empirical methods used \\ over time?} & 0.8245 & yes \\ \hline
re & \makecell[l]{What are the typical \\ outcomes of empirical RE \\ studies on process maturity?} & \makecell[l]{Which sub-fields of SE \\ and RE do the empirical \\ studies cover over time?} & 0.4718 & no \\ \hline
\end{tabular}
\end{table}
Precision measures the accuracy of the CQs generated by the LLM. 
It is the ratio of true positives to the total number of CQs generated by the LLM.
%Recall measures the completeness of the CQs generated by the LLM.
%It is the ratio of true positives to the total number of ground truth CQs.
The precision of the CQs generated by the LLM can be defined as follows:
\begin{equation}
\text{Precision} = \frac{TP}{TP + FP}
\end{equation}
True positives (TP) is the number of valid $CQ_{gen}$ and false positives (FP) is the number of invalid $CQ_{gen}$.

% \begin{description}
%     \item True Positives (TP): The number of valid $CQ_{gen}$
%     \item False Positives (FP): The number of invalid $CQ_{gen}$
%     %\item False Negatives (FN): The number of $CQ_{gt}$ that do not have a valid $CQ_{gen}$
% \end{description}
% The recall of the CQs generated by the LLM can be defined as:
% \begin{equation}
% \text{Recall} = \frac{TP}{TP + FN}
% \end{equation}
Table \ref{table:cqs} shows some examples of the $CQ_{gen}$ and their matched $CQ_{gt}$ with cosine similarity scores for two domain engineering tasks.

The consistency in this task refers to how similar or stable the generated CQs are across multiple runs for each temperature setting.
we use task performance variance and text similarity variance to measure the consistency of the LLM's output across different temperatures.
The variance in task performance is defined as the standard deviation of precision for all iterations in each temperature setting, denoted as $std_{precision}$.
The variance in text similarity is defined as the standard deviation of cosine similarity of the generate text across different temperature, denoted as $std_{cosine}$
Since we run the experiments with each setting for 10 times, we take the average of each metric.

Additionally, we perform an Analysis of Variance (ANOVA) test to assess the significance of $N_{paper}$ and $temp$ on the task performance.

\section{Results and discussion}
\label{sec:results}
\subsection{Performance of generating CQs}
\vspace{-7mm}
\begin{figure}[h]
    \centering
    \includegraphics[width=1.0\textwidth]{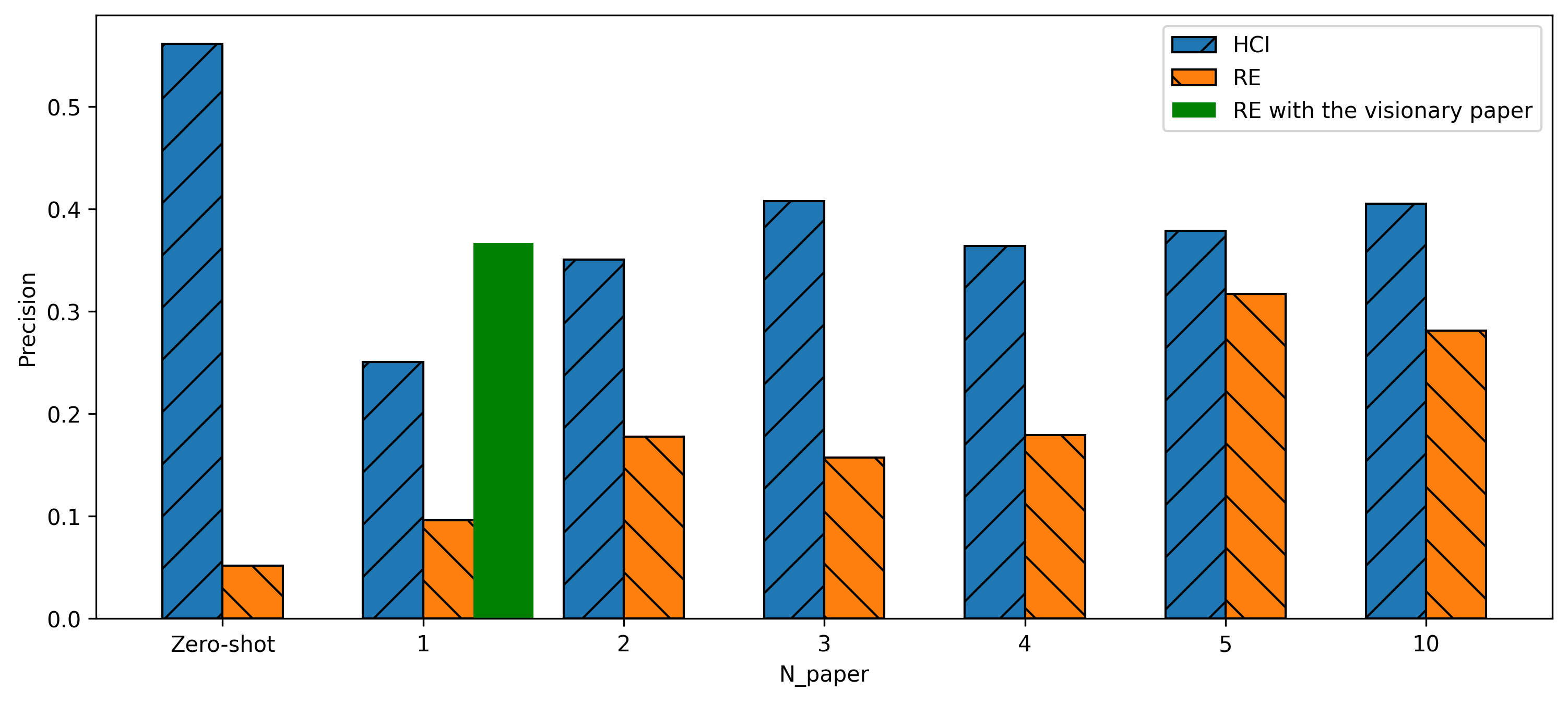}
    \caption{Precision of using RAG with gpt-4o to generate CQs, compared to zero-shot prompting}
    \label{fig:precision}
\end{figure}
\vspace{-5mm}
Figure \ref{fig:precision} shows the precision of our RAG approach, compared to zero-shot prompting in generating CQs for two domain ontology engineering tasks, with different number of papers in the knowledge base for the RAG.

As a first look, we observe that for the task in the RE domain, our RAG approach performs much better than zero-shot prompting. 
%(see the orange bar in Figure \ref{fig:precision}).
As $N_{paper}$ increased, there is a noticeable trend where precision increase marginally.
These suggest that the addition of domain knowledge could enhance the understanding capacity of LLMs for generating CQs for domain ontology engineering.
In particular, the best precision is achieved when we use the only visionary paper as input to the knowledge base in the RAG pipeline, 
%(see the green bar in Figure \ref{fig:precision}), 
following the methodology of the KG-EmpiRE paper to identify ground-truth CQs.
These indicate that with a good selection of papers in the knowledge base, small size of a knowledge base could achieve good performance, which is very important from the perspective of token cost in closed-source LLMs such as OpenAI's GPT models.
%Moreover, as we increase the size of the knowledge base with $N_{paper}$ ranging from 2 to 10, the precision marginally increased, which illustrates the intricate balance between the breadth of information and the maintenance of high data quality in improving the task performance.

For the precision of the task in the HCI domain, 
%(see the blue bar in Figure \ref{fig:precision})
zero-shot prompting yields the highest precision, marginally better than our RAG approach across all $N_{paper}$.
Similar to the performance of the task in the RE domain, increasing the number of papers $N_{paper}$ ranging from 1 to 10 generally improves the precision, but does not surpass the precision achieved with zero-shot prompting.
Therefore, taking into consideration of the precision performance and the token cost for processing the documents in knowledge base, our RAG approach might not fit for generating CQs for HCIO, compared to zero-shot prompting.

From the perspective of zero-shot prompting, we observe that the precision in HCI is much higher than the precision in RE, which we think is due to the degree of abstraction of the target ontology or knowledge graph. 
The more concrete the ontology/KG is, the more domain knowledge is required for LLMs to perform well in the tasks of generating CQs for ontology engineering.
For instance, the purpose of the HCIO is to develop a reference ontology to represent the core concepts in HCI, while the purpose of the KG-EmpiRE is to construct a knowledge graph to capture the state and evolution in RE.
More domain knowledge such as specific methods used in RE is required to construct the KG-EmpiRE so as to answer the example $CQ_{gt}$ shown in Table~\ref{table:cqs}, \textit{How often are which empirical methods used over time?}.
Therefore, our RAG approach outperforms zero-shot prompting in generating CQs for KG-EmpiRE, while zero-shot prompting outperforms the proposed RAG approach in generating CQs for HCIO.

In general, our RAG approach works well in the task of generating CQs for more concrete ontologies/KGs, compared to zero-shot prompting.
Moreover, increasing the number of papers $N_{paper}$ in the RAG pipeline generally improves the precision of generating CQs for domain ontologies/KGs.

The ANOVA test results also confirm that $N_{paper}$ has a significant impact on the task performance ($p<0.001$).

\subsection{Consistency of LLMs}
\begin{figure}[h]
    \centering
    \includegraphics[width=.5\textwidth]{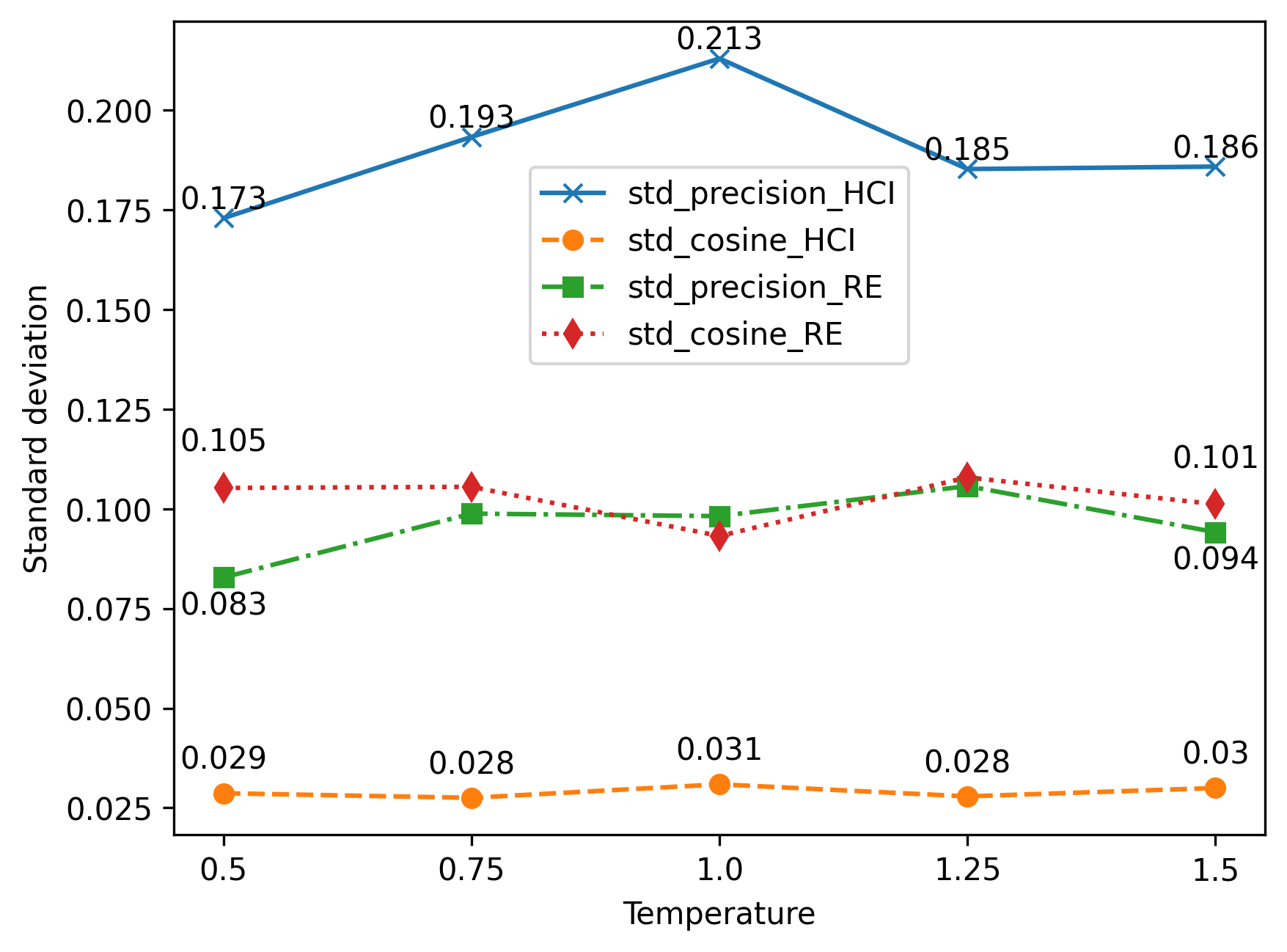}
    \caption{Standard deviation of precision for the task performance and standard deviation of cosine similarity for generated text with different temperature.}
    \label{fig:std}
\end{figure}

Figure \ref{fig:std} shows the standard deviation of precision for task performance and the standard deviation of cosine similarity for the generated text with different temperature settings in two domain ontology engineering tasks.

Apparently, there are no obvious patterns about how consistency changes by different temperatures regardless of different consistency metrics, which contrasts with the assumption that with higher temperature, less probable tokens are more likely to be sampled, resulting in more random outputs.
This suggests that the temperature setting would not affect the task performance of generating CQs in our RAG pipeline.

The ANOVA test results also confirm that $temp$ doesn't have a significant impact on the task performance ($p>0.05$).

From the perspective of different domains, the vertical distance between $std_{precision}$ and $std_{cosine}$ in RE is significantly smaller than that observed in HCI.
This suggests a reduced difference among two standard deviation for RE as opposed to HCI across the temperature range.
This phenomenon may be attributed to the quantity of $CQ_{gt}$.
Specifically, with 77 $CQ_{gt}$ in RE, there is a higher likelihood that the 77 $CQ_{gt}$ generated over 10 iterations exhibit overlap.

\section{Conclusion}
\label{sec:conclusion}
This study aimed to examine the capabilities of retrieval-augmented generation (RAG)-based large language models (LLMs) to generate competency questions (CQs) for ontology engineering. 
Additionally, we explored how various hyperparameters within the RAG pipeline influence the task performance. 
A prompt template with four variables was specifically designed to direct LLMs in generating CQs for domain ontologies and knowledge graphs.
Using one LLM (gpt-4o), two LLMs techniques (zero-shot prompting and RAG) and five temperature settings, we replicated two expert-driven ontology engineering tasks from different domains.
Experimental results revealed that our RAG approach works well in the task of generating CQs for concrete ontologies/KGs that required more domain knowledge, such as generating CQs for KG-EmpiRE, compared to zero-shot prompting.
Moreover, increasing the number of papers $N_{paper}$ in the RAG pipeline generally improves the precision of generating CQs for domain ontologies/KG.
It is interesting to note that the temperature setting does not have a significant impact on our approach.

Our RAG approach accelerates ontology engineering by automatically generating CQs.
The generated CQs can serve as candidate CQs for domain experts in the design phase of ontology engineering and can also be utilized to evaluate existing ontologies and knowledge graphs.
Future work will focus on using our RAG pipeline for generating CQs in more domain ontology engineering tasks to investigate the generalization of our approach.
Furthermore, due to the token cost of OpenAI's GPT models, we would like to explore the capability of open-source LLMs, such as Meta's Llama and Google's Gemma for generating CQs.

\bibliographystyle{splncs04}
\bibliography{ref}

\end{document}